\documentclass[cameraready]{Interspeech}

\title{Multilingual Phonological Feature Recognition with \\
Self-Supervised Speech Models}

\author[affiliation={1}]{Abner}{Hernandez}
\author[affiliation={1,2}]{Tom\'as}{Arias-Vergara}
\author[affiliation={1}]{Daiqi}{Liu}
\author[affiliation={1}]{Andreas}{Maier}
\makeatletter
\protected@edef\authorlist{\authorlist\\}
\renewcommand{\authorsep}{}
\makeatother
\author[affiliation={1,2}]{Paula~Andrea}{P\'erez-Toro}

\address{
    $^1$ Pattern Recognition Lab, Friedrich-Alexander-Universität Erlangen-Nürnberg, Germany \\
    $^2$ GITA Lab. Facultad de Ingeniería. Universidad de Antioquia UdeA, Medellín, Colombia
}

\email{abner.hernandez@fau.de}


\usepackage{comment}
\usepackage{graphicx}
\usepackage[table]{xcolor}
\usepackage{tabularx}
\usepackage{float}
\usepackage{stfloats}
\usepackage{multirow}
\usepackage{tipa}
\usepackage{cleveref}

\begin{document}

\maketitle

\begin{abstract}
Phonological features provide a language-general and linguistically grounded representation of speech. We present PhonoQ-2.0, a multilingual frame-level phonological feature recognizer built on self-supervised speech models. The system directly predicts a structured 22-dimensional feature vector per frame encoding manner, vowel quality, place, and voicing, instead of deriving features from phoneme outputs. To ensure phonologically coherent predictions, we introduce a manner-conditioned gating mechanism that activates valid feature groups. Evaluated across multiple languages and corpora, PhonoQ-2.0 achieves an average macro-F1 of 91.3\% in-domain and 88.9\% out-of-domain. Compared to a strong CTC phoneme baseline, it delivers consistent gains of +8.8 F1 in-domain and +8.6 out-of-domain on average. In unseen-language evaluation, PhonoQ-2.0 improves macro-F1 from 66.9\% to 73.6\% (+6.7 on average), with gains of up to +10.8 points.

\end{abstract}

\section{Introduction}

Self-supervised learning (SSL)-based speech models have made phoneme recognition highly accurate across many languages~\cite{conneau21_interspeech}. Most downstream systems therefore focus on phoneme prediction or end-to-end word modeling. In such pipelines, phonological information is derived indirectly: models predict phoneme sequences, and phonological features are obtained via deterministic phone-to-feature mapping~\cite{mortensen2016panphon}. This phoneme-first strategy does not explicitly model phonological structure. Moreover, recent work shows that SSL representations encode articulatory and phonological structure~\cite{choi2025leveraging,cho2024self}, suggesting that phonological abstractions are already present in pretrained models.

Phonological features characterize speech through articulatory properties such as manner, place of articulation, vowel height and backness, and voicing, forming a language-general representation~\cite{clements1995internal}. 

Prior work has explored direct phonological feature modeling from acoustic input. Neural frame-level detectors have demonstrated that distinctive feature posteriors can be estimated directly from speech~\cite{vasquezcorrea19_interspeech,tadavarthy24_interspeech}. Phonological features have also been incorporated into ASR systems, for example through phonological bottleneck layers for pathological speech analysis~\cite{thienpondt2025weakly}, or as auxiliary IPA and articulatory targets to improve multilingual generalization~\cite{lee2025leveraging,kim2025improving}. 
In clinical contexts, phonological features have been used in discriminative models for dysarthric speech~\cite{rudzicz2009phonological}, and more recently for assessing phonological precision and contrast realization~\cite{arias2023measuring,arias2024contrastive,liu2025audio}.

However, most prior systems are monolingual, treat features as independent labels, or use them only as auxiliary targets, and do not directly compare structured phonological prediction against strong phoneme-based baselines under multilingual and cross-domain conditions.


We base our approach on the insight that explicitly modeling phonological structure can improve robustness and cross-lingual generalization beyond phoneme prediction. Building on the concept of phonological class prediction exemplified by PhonoQ~\cite{arias2022analysis}, we introduce \textbf{PhonoQ-2.0},\footnote{Code, trained models, and usage instructions are available at: \url{https://github.com/abnerLing/PhonoQ-2.0}} a multilingual frame-level phonological feature recognizer trained jointly across languages and evaluated under both domain and language shift.

In contrast to the original PhonoQ architecture, which relied on convolutional and recurrent layers trained from scratch and predicted phonological classes through a single shared output layer, PhonoQ-2.0 leverages a pretrained multilingual self-supervised speech encoder and adopts a structured multi-head design. The model predicts a 22-dimensional phonological feature vector per frame using dedicated heads for manner, vowel height and backness, place of articulation, and voicing. A manner-conditioned gating mechanism restricts vowel and place predictions to compatible manner classes, explicitly enforcing phonological coherence and reflecting the internal structure of phonological systems.

Our contributions are:
\begin{itemize}
    \item \textbf{Multilingual Phonological Feature Modeling:} A multilingual phonological recognizer that directly predicts a structured feature inventory (manner, vowel, place, voicing) from speech, rather than deriving features from phoneme outputs.
    \item \textbf{Phonologically coherent decoding:} We introduce a manner-conditioned gating mechanism that enforces valid feature group activation (e.g., vowel features only for vowels), yielding interpretable and internally consistent predictions.
    \item \textbf{Systematic multilingual evaluation:} We benchmark in-domain, cross-corpus out-of-domain, and unseen-language transfer, and compare against a strong CTC phoneme baseline under matched SSL backbones in the same phonological feature space.
\end{itemize}



\section{Method}

\subsection{Phonological Feature Representation}

\begin{table}[t]
\centering
\caption{Phonological feature inventory representing as a 22-dimensional vector spanning four feature groups.}
\footnotesize
\renewcommand{\arraystretch}{1.0}
\begin{tabularx}{\columnwidth}{@{}>{\centering\arraybackslash}l>{\centering\arraybackslash}l>{\centering\arraybackslash}X@{}}
\toprule
Group & Class & Examples (IPA) \\
\midrule
\multirow{9}{*}{Manner}        & Silence      & -- \\
                               & Stop         & \textipa{/p t k b d g/} \\
                               & Nasal        & \textipa{/m n N/} \\
                               & Rhotic       & \textipa{/r \;R/} \\
                               & Fricative    & \textipa{/f v s z S x/} \\
                               & Affricate    & \textipa{/ts tS pf/} \\
                               & Approximant  & \textipa{/j w/} \\
                               & Lateral      & \textipa{/l \;L/} \\
                               & Vowel        & \textipa{/a e i o u/} \\
\midrule
\multirow{3}{*}{\shortstack{Vowel\\Height}}   & High    & \textipa{/i u I \textupsilon/} \\
                                              & Mid     & \textipa{/e o \textschwa\ E/} \\
                                              & Low     & \textipa{/a \ae/} \\
\midrule
\multirow{3}{*}{\shortstack{Vowel\\Backness}} & Front   & \textipa{/i e E y \o/} \\
                                              & Central & \textipa{/\textschwa\ a/} \\
                                              & Back    & \textipa{/u o \textupsilon/} \\
\midrule
\multirow{5}{*}{Place}         & Labial       & \textipa{/p b m f v/} \\
                               & Alveolar     & \textipa{/t d n s z l/} \\
                               & Velar        & \textipa{/k g N x/} \\
                               & Palatal      & \textipa{/j c \textctj \textltailn/} \\
                               & Postalveolar & \textipa{/S Z tS dZ/} \\
\midrule
\multirow{2}{*}{Voicing}       & Voiceless    & \textipa{/p t k f s S/} \\
                               & Voiced       & \textipa{/b d g v z m/} \\
\bottomrule
\end{tabularx}
\label{tab:phonoq_features}
\end{table}
We compare PhonoQ-2.0, a structured multi-head phonological recognizer that predicts features directly, and \textbf{CTC-Phoneme}, a phoneme recognizer that derives phonological classes via post-hoc mapping. Both systems share the same pretrained acoustic backbone. PhonoQ-2.0 represents each frame as a 22-dimensional phonological vector decomposed into four groups: manner (9 classes), vowel height and backness (6 classes), place (5 classes), and voicing (2 classes), as detailed in \Cref{tab:phonoq_features}.

Phone labels from Montreal Forced Aligner (MFA)~\cite{mcauliffe17_interspeech} alignments are canonicalized before feature mapping. Suprasegmental markers (stress, length, nasalization) and affricate ligatures (e.g., tie-bar variants of \textipa{tS}, \textipa{dZ}) are normalized across all languages. Language-specific steps include collapsing aspirated stops (\textipa{p\super{h}}, \textipa{t\super{h}}, \textipa{k\super{h}} $\to$ \textipa{p}, \textipa{t}, \textipa{k}) and syllabic consonants (\textipa{\s{l}}, \textipa{\s{n}}) in German, reducing the alveolar tap (\textipa{\textfishhookr} $\to$ \textipa{\textturnr}) and rhotic vowels (\textipa{\textrhookschwa}, \textipa{\textrhookrevepsilon} $\to$ \textipa{\textschwa}) in English, and retaining Spanish intervocalic approximant allophones \textipa{(B, D, G)} as fricatives to reflect their surface realization.

Diphthongs are treated as vowels with height and backness assigned from the nucleus. For example, \textipa{aI}/\textipa{aU} map to low-central across German and English. 
PhonoQ-2.0 predicts the 22 dimensions via a structured multi-head design: the manner head determines silence, consonant, or vowel, and vowel features (height, backness) and place are only predicted for the relevant manner class, enforcing phonologically coherent outputs.

\subsection{Model Architecture}

\begin{figure}
    \centering
    \includegraphics[width=\linewidth]{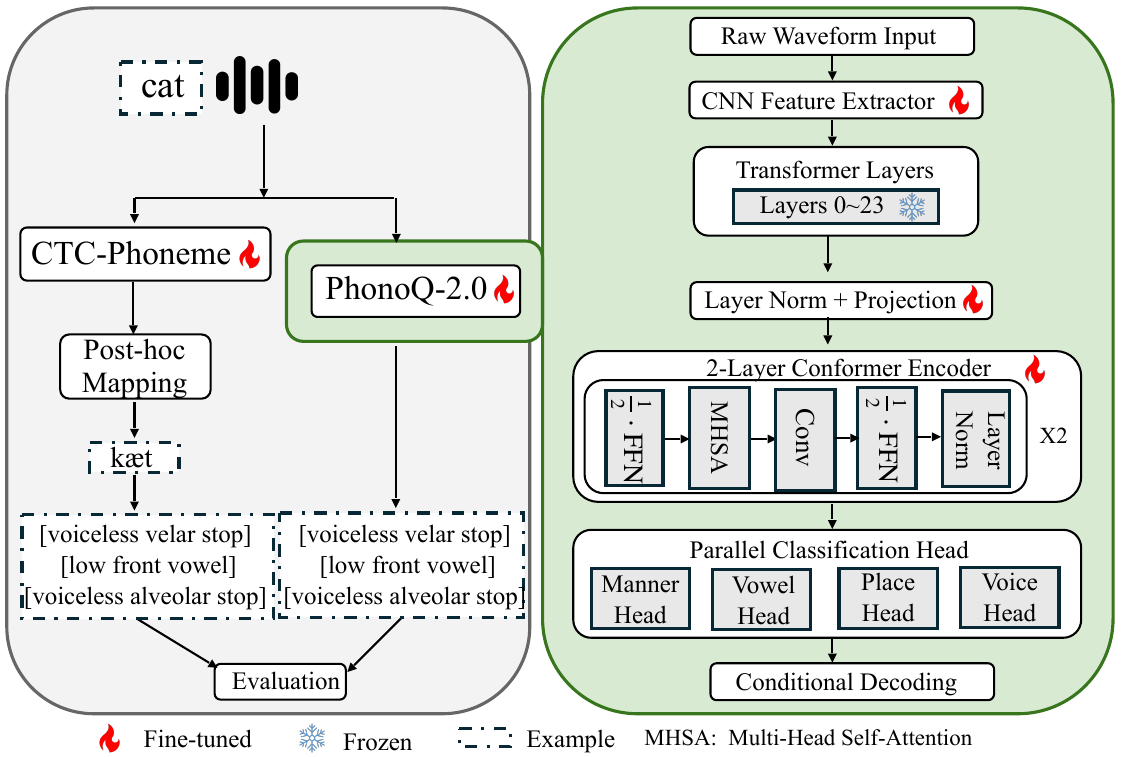}
    \caption{Pipeline comparison. Left: CTC-Phoneme predicts phonemes, which are mapped to phonological classes, while PhonoQ-2.0 predicts phonological features directly. Right: PhonoQ-2.0 uses a pretrained XLSR-ft encoder followed by a 2-layer Conformer and structured classification heads with conditional decoding enforcing valid feature combinations.}
    \label{fig:methodology}
\end{figure}

All systems use XLSR-ft,\footnote{\texttt{facebook/wav2vec2-xlsr-53-espeak-cv-ft}} a multilingual wav2vec~2.0 model fine-tuned for phoneme recognition~\cite{xu2021simple}, as the acoustic backbone, with all 24 Transformer layers frozen during fine-tuning (\Cref{fig:methodology}). PhonoQ-2.0 applies a shared linear projection followed by a 2-layer Conformer ($d=512$, 4 attention heads) with relative position bias, on top of which four cross-entropy-trained heads predict manner (9-way), vowel features (6-way), place (5-way), and voicing (2-way). The key distinction from a conventional multi-label classifier is therefore not the use of additional output heads alone, but the explicit modeling of dependencies among phonological feature groups through manner-conditioned decoding.

\subsection{Phonetic Baseline with Class Mapping}

We fine-tune the same pretrained XLSR-ft backbone with a CTC phoneme prediction head on MFA-aligned phone sequences.
To enable comparison in the same representation space, predicted phonemes are deterministically mapped to the 22-dimensional PhonoQ-2.0 feature inventory using a fixed phone-to-feature lookup table shared across languages. Although PhonoQ-2.0 is trained with frame-level labels derived from MFA alignments whereas CTC-Phoneme learns phone sequences implicitly, both systems are evaluated over the same MFA-derived phone segments after mapping predictions to the shared 22-dimensional phonological feature space.

\section{Experimental Setup}

\subsection{Datasets}

We train and evaluate on four languages spanning three families: Germanic (English, German), Romance (Spanish), and Slavic (Czech).
For \textbf{Spanish}, training data combine 14.5h of CommonPhone (CP)~\cite{klumpp2022common} and a Latin American Spanish corpus comprising publicly available speech from Argentina, Chile, Colombia, Peru, Puerto Rico, and Venezuela (38h)~\cite{guevara2020crowdsourcing}. \textbf{English} training data include CP (14.1h), TIMIT (4h)~\cite{garofolo1993darpa}, and LibriSpeech (36h)~\cite{panayotov2015librispeech}. \textbf{German} data combine CP (13.6h) and the Carina corpus (40h)~\cite{kath2022carina}. As CP does not include \textbf{Czech}, we construct an analogous split using CommonVoice and ParlaSpeech~\cite{ardila2020common,parlaspeech,parlaspeech3}, contributing 16h and 40h respectively.

Each language comprises approximately 52 to 56 hours of training speech. Development (dev) and test sets contain approximately 2 to 3 hours per language. Development uses the CP dev partition (or a matched CommonVoice dev split for Czech). Testing is performed on CP test splits, as well as cross-corpus evaluation on the official test splits of FLEURS~\cite{fleurs2022arxiv} (EN: 51min, ES: 53min, DE: 52min, CZ: 51min) and VoxPopuli~\cite{wang-etal-2021-voxpopuli} (EN: 3h10min, ES: 2h32min, DE: 3h15min, CZ: 2h3min). PhonoQ-2.0 and CTC-Phoneme are trained on identical language-specific data and evaluated on the same splits.

\subsection{Training Configuration}

PhonoQ-2.0 is optimized using AdamW with weight decay 0.01. 
We use separate learning rates of $1\times10^{-5}$ for the encoder and $1\times10^{-3}$ for the prediction heads. Training runs for up to 40 epochs with a batch size of 16 and gradient clipping set to 0.5. We apply label smoothing (0.05), center-frame label alignment, and per-head class weighting to mitigate class imbalance. Early stopping is based on development macro-F1.

CTC-Phoneme is trained using the HuggingFace Trainer with default settings. We use a learning rate of $3\times10^{-5}$ and weight decay 0.01, and train for 10 epochs with a batch size of 8. Mixed-precision (fp16) training is applied, and model selection is based on development phone error rate (PER).

\subsection{Evaluation Protocol}

PhonoQ-2.0 is trained to predict structured phonological feature vectors at the frame level using MFA-aligned TextGrids converted to 50\,fps representations. Although supervision is frame-aligned, all results reported in this paper are computed at the phone segment level to enable direct comparison with the CTC-Phoneme baseline. For PhonoQ-2.0, frame-level logits are aggregated within each non-silence phone interval by summing over frames. A single segment-level prediction is then obtained via head-wise argmax, yielding a structured 22-dimensional phonological vector per segment.

CTC-Phoneme predicts phoneme sequences, which are deterministically mapped to the same phonological feature vectors. By construction, these features are constant within each phone segment. Segment-level evaluation reports macro-F1, per-feature F1, and grouped F1 (manner, height, backness, place, voice), excluding silence.

\section{Results}

We evaluate PhonoQ-2.0 and CTC-Phoneme under three increasingly challenging conditions: \textbf{(1) In-domain} (CP and CV test splits), \textbf{(2) Out-of-domain} (FLEURS, VoxPopuli), and \textbf{(3) Zero-shot cross-lingual} (French, Italian, Russian from CP, unseen during training).
All evaluations are conducted at the phone-segment level within the shared phonological feature space. Performance is reported as macro-F1 over the 21 non-silence features, with PER also reported for CTC-Phoneme.

\subsection{In-Domain Evaluation}

Table~\ref{tab:indomain_ood} reports segment-level phonological macro-F1 on the
CommonVoice test sets for all four languages. The CTC-Phoneme system achieves competitive
phoneme recognition (6.80\% PER overall), yet its macro-F1 when mapped to the
phonological feature space reaches only 82.5\%, substantially below PhonoQ-2.0 at
91.3\%. Improvements are consistent across languages, ranging from +7.3 F1 in Czech to
+11.6 F1 in English. Notably, English exhibits both the highest PER (9.52\%) and the
largest phonological gain, suggesting that explicit structured modeling is particularly
beneficial under more challenging phonetic conditions.

\begin{table}[t]
\centering
\caption{In-domain and out-of-domain evaluation. PER (\%,~$\downarrow$; CTC-Ph only) and segment-level F1 (\%,~$\uparrow$). Vowel~=~average of height and backness. Avg~=~macro-F1 over 21 non-silence features.}
\label{tab:indomain_ood}
\renewcommand{\arraystretch}{0.95}
\resizebox{\columnwidth}{!}{
\begin{tabular}{@{}ll l@{\hspace{6pt}}cccccc@{}}
\toprule
Lang. & Corpus & System & PER & Man. & Vow. & Pl. & Voi. & Avg \\
\midrule

\multirow{6}{*}{Czech}
 & \multirow{2}{*}{CP$^\dagger$}
    & \cellcolor{gray!15}CTC-Ph & \cellcolor{gray!15}8.68 & \cellcolor{gray!15}80.3 & \cellcolor{gray!15}84.5 & \cellcolor{gray!15}85.7 & \cellcolor{gray!15}93.2 & \cellcolor{gray!15}84.0 \\
 &  & PhonoQ-2.0 & -- & 90.9 & 90.5 & 90.4 & 97.1 & 91.3 \\
 & \multirow{2}{*}{FLEURS$^\star$}
    & \cellcolor{gray!15}CTC-Ph & \cellcolor{gray!15}17.7 & \cellcolor{gray!15}76.9 & \cellcolor{gray!15}78.4 & \cellcolor{gray!15}83.3 & \cellcolor{gray!15}92.5 & \cellcolor{gray!15}80.4 \\
 &  & PhonoQ-2.0 & -- & 89.2 & 87.6 & 90.0 & 97.2 & 89.7 \\
 & \multirow{2}{*}{VoxPopuli$^\star$}
    & \cellcolor{gray!15}CTC-Ph & \cellcolor{gray!15}7.4 & \cellcolor{gray!15}81.4 & \cellcolor{gray!15}87.3 & \cellcolor{gray!15}87.1 & \cellcolor{gray!15}93.2 & \cellcolor{gray!15}85.6 \\
 &  & PhonoQ-2.0 & -- & 91.8 & 92.8 & 91.5 & 97.4 & 92.5 \\
\midrule

\multirow{6}{*}{German}
 & \multirow{2}{*}{CP$^\dagger$}
    & \cellcolor{gray!15}CTC-Ph & \cellcolor{gray!15}5.50 & \cellcolor{gray!15}77.9 & \cellcolor{gray!15}81.4 & \cellcolor{gray!15}84.9 & \cellcolor{gray!15}91.2 & \cellcolor{gray!15}81.8 \\
 &  & PhonoQ-2.0 & -- & 88.9 & 90.4 & 91.3 & 96.9 & 90.6 \\
 & \multirow{2}{*}{FLEURS$^\star$}
    & \cellcolor{gray!15}CTC-Ph & \cellcolor{gray!15}9.1 & \cellcolor{gray!15}75.8 & \cellcolor{gray!15}78.2 & \cellcolor{gray!15}83.2 & \cellcolor{gray!15}90.6 & \cellcolor{gray!15}79.7 \\
 &  & PhonoQ-2.0 & -- & 86.0 & 86.5 & 88.2 & 95.4 & 87.6 \\
 & \multirow{2}{*}{VoxPopuli$^\star$}
    & \cellcolor{gray!15}CTC-Ph & \cellcolor{gray!15}12.9 & \cellcolor{gray!15}71.6 & \cellcolor{gray!15}74.3 & \cellcolor{gray!15}79.6 & \cellcolor{gray!15}88.8 & \cellcolor{gray!15}75.9 \\
 &  & PhonoQ-2.0 & -- & 80.6 & 82.2 & 82.6 & 92.0 & 82.6 \\
\midrule

\multirow{6}{*}{English}
 & \multirow{2}{*}{CP$^\dagger$}
    & \cellcolor{gray!15}CTC-Ph & \cellcolor{gray!15}9.52 & \cellcolor{gray!15}77.4 & \cellcolor{gray!15}75.9 & \cellcolor{gray!15}77.8 & \cellcolor{gray!15}91.4 & \cellcolor{gray!15}78.4 \\
 &  & PhonoQ-2.0 & -- & 90.5 & 87.2 & 89.9 & 96.1 & 90.0 \\
 & \multirow{2}{*}{FLEURS$^\star$}
    & \cellcolor{gray!15}CTC-Ph & \cellcolor{gray!15}12.1 & \cellcolor{gray!15}76.6 & \cellcolor{gray!15}75.9 & \cellcolor{gray!15}78.1 & \cellcolor{gray!15}91.0 & \cellcolor{gray!15}78.1 \\
 &  & PhonoQ-2.0 & -- & 89.1 & 87.3 & 87.4 & 94.6 & 88.7 \\
 & \multirow{2}{*}{VoxPopuli$^\star$}
    & \cellcolor{gray!15}CTC-Ph & \cellcolor{gray!15}16.7 & \cellcolor{gray!15}76.1 & \cellcolor{gray!15}72.9 & \cellcolor{gray!15}78.0 & \cellcolor{gray!15}90.5 & \cellcolor{gray!15}77.0 \\
 &  & PhonoQ-2.0 & -- & 86.3 & 81.0 & 84.4 & 94.6 & 85.1 \\
\midrule

\multirow{6}{*}{Spanish}
 & \multirow{2}{*}{CP$^\dagger$}
    & \cellcolor{gray!15}CTC-Ph & \cellcolor{gray!15}3.49 & \cellcolor{gray!15}83.6 & \cellcolor{gray!15}90.1 & \cellcolor{gray!15}80.5 & \cellcolor{gray!15}94.6 & \cellcolor{gray!15}85.7 \\
 &  & PhonoQ-2.0 & -- & 92.6 & 93.4 & 92.3 & 97.7 & 93.3 \\
 & \multirow{2}{*}{FLEURS$^\star$}
    & \cellcolor{gray!15}CTC-Ph & \cellcolor{gray!15}5.8 & \cellcolor{gray!15}82.1 & \cellcolor{gray!15}89.3 & \cellcolor{gray!15}76.9 & \cellcolor{gray!15}93.9 & \cellcolor{gray!15}84.0 \\
 &  & PhonoQ-2.0 & -- & 93.2 & 93.7 & 92.3 & 97.8 & 93.6 \\
 & \multirow{2}{*}{VoxPopuli$^\star$}
    & \cellcolor{gray!15}CTC-Ph & \cellcolor{gray!15}7.4 & \cellcolor{gray!15}78.8 & \cellcolor{gray!15}85.8 & \cellcolor{gray!15}76.2 & \cellcolor{gray!15}93.2 & \cellcolor{gray!15}81.5 \\
 &  & PhonoQ-2.0 & -- & 90.3 & 90.4 & 91.3 & 96.2 & 91.1 \\
\bottomrule
\multicolumn{9}{l}{$^\dagger$ In-domain (CP). $^\star$ Out-of-domain (FLEURS, VoxPopuli). Man.: Manner,}\\
\multicolumn{9}{l}{Vow.: Vowel, Pl.: Place, Voi.: Voicing, CTC-Ph: CTC-Phoneme, Lang.: Language.}
\end{tabular}
}
\end{table}

\begin{figure*}[t]
    \centering
    \includegraphics[width=0.88\textwidth]{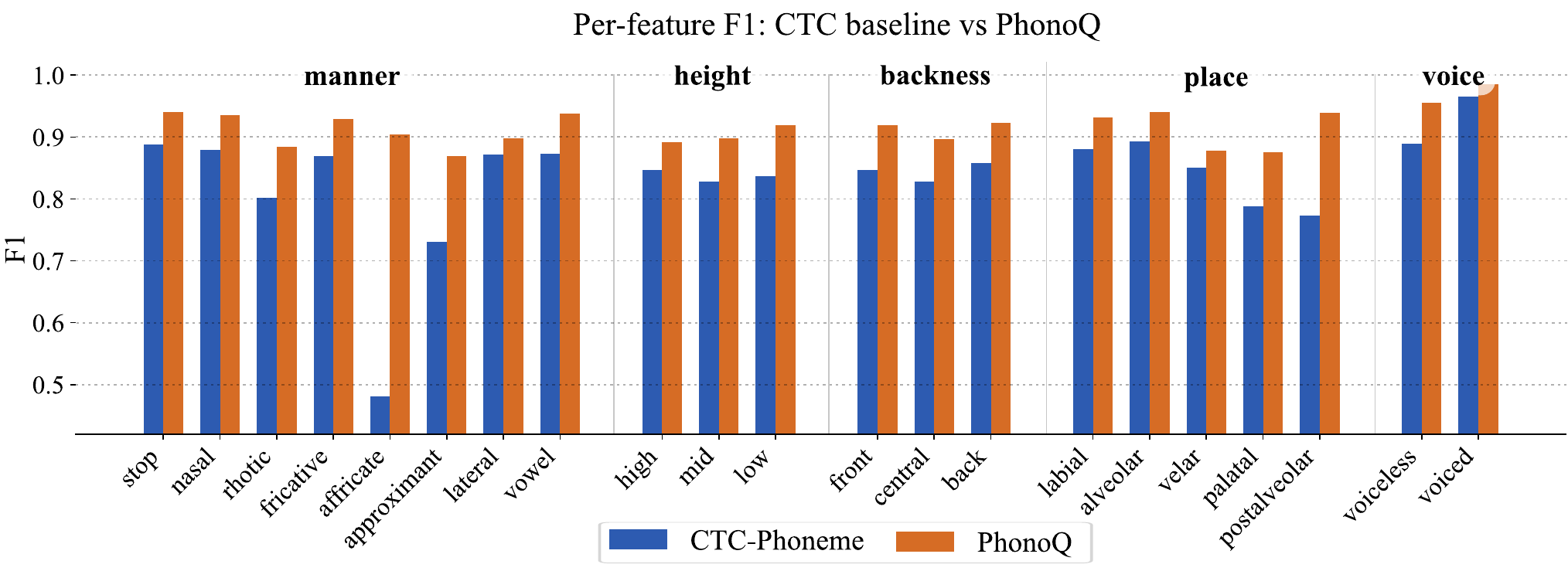}
    \caption{Per-feature segment-level F1 (\%) on the in-domain CV-test set for CTC-Phoneme and PhonoQ-2.0.}
    \label{fig:per_feature_f1}
\end{figure*}

\begin{figure}[t]
    \centering
    \includegraphics[width=\columnwidth]{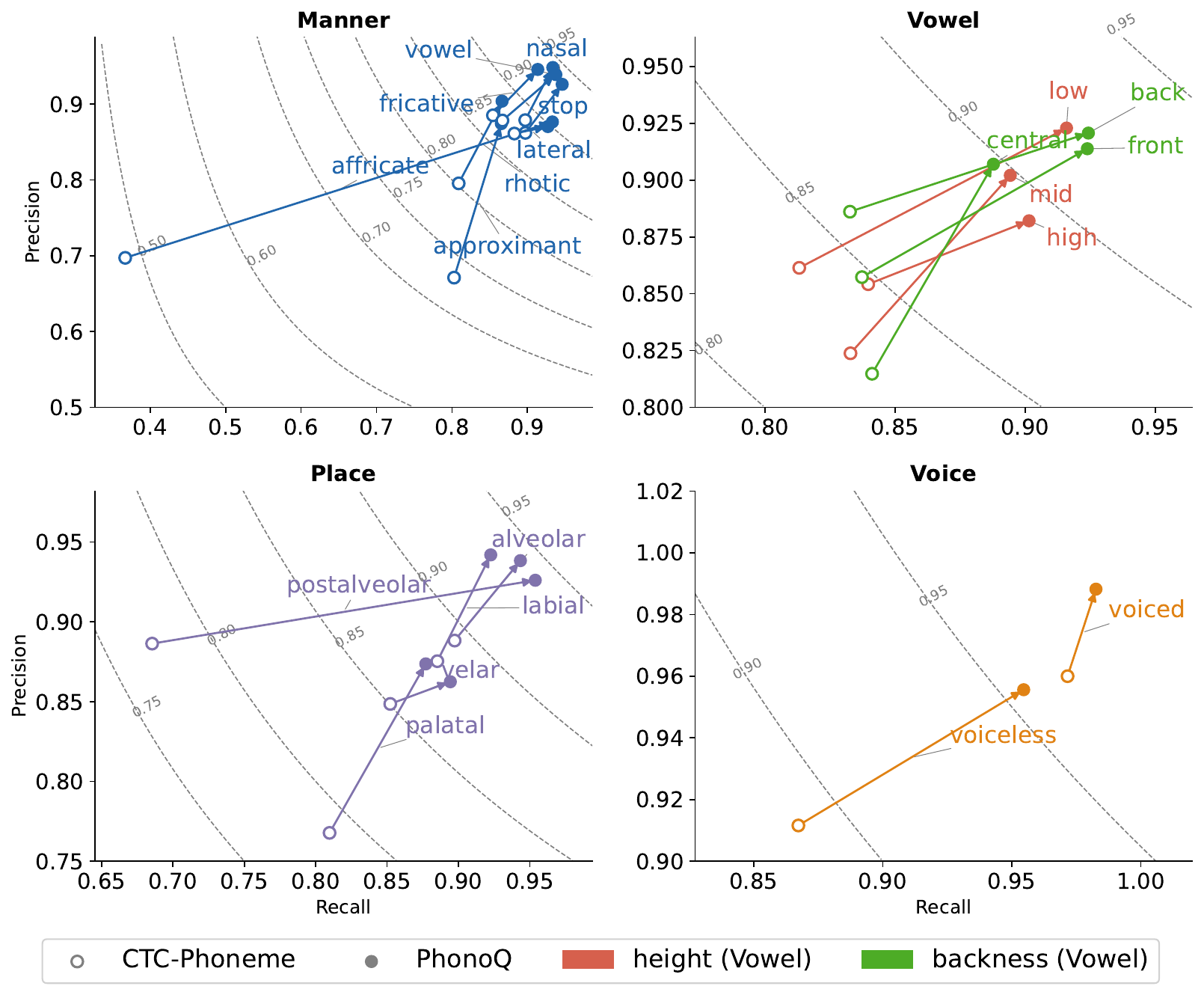}
    \caption{Precision and recall per phonological feature on the in-domain CV-test. Arrows connect CTC-Phoneme to PhonoQ-2.0.}
    \label{fig:per_feature_f1_scatter}
    \vspace{-4mm}
\end{figure}

\subsection{Out-of-Domain Evaluation}

Out-of-domain performance on FLEURS and VoxPopuli (Table~\ref{tab:indomain_ood}) follows the expected pattern: CTC-Phoneme PER rises to 11.18\% and 11.10\% respectively, with corresponding drops in macro-F1 for both systems. Despite this degradation, PhonoQ-2.0 maintains a substantial and consistent advantage across all languages and corpora, +9.3 F1 on FLEURS (89.9\% vs.\ 80.6\%) and +7.8 F1 on VoxPopuli (87.8\% vs.\ 80.0\%).

Table~\ref{tab:phonoq_v1_v2_comparison} compares PhonoQ-2.0 against its predecessor PhonoQ~\cite{arias2022analysis} on German and Spanish, the two languages supported by both systems. To enable a fair comparison, evaluation is restricted to the 12 phonological dimensions shared between both systems: six manner classes (stop, nasal, rhotic, fricative, lateral, vowel), four place classes (labial, alveolar, velar, postalveolar), and voicing. Rhotics are mapped to PhonoQ's trill class for this comparison. Macro-F1 is computed over these 12 shared dimensions only.

\begin{table}[t]
\centering
\caption{Comparison of original PhonoQ and PhonoQ-2.0 on FLEURS and VoxPopuli (DE, ES). Group F1 (\%,~$\uparrow$) and macro F1 restricted to 12 shared dimensions.}
\label{tab:phonoq_v1_v2_comparison}
\resizebox{\columnwidth}{!}{
\begin{tabular}{@{}lll rrrr@{}}
\toprule
Lang. & Corpus & System & Man. & Pl. & Voi. & Macro \\
\midrule
\multirow{4}{*}{German} & \multirow{2}{*}{FLEURS} & PhonoQ & 68.6 & 66.3 & 85.5 & 70.7 \\
 &  & PhonoQ-2.0 & \textbf{86.0} & \textbf{88.2} & \textbf{95.4} & \textbf{88.5} \\
\cmidrule(l){2-7}
 & \multirow{2}{*}{VoxPopuli} & PhonoQ & 62.2 & 59.2 & 84.8 & 64.9 \\
 &  & PhonoQ-2.0 & \textbf{80.6} & \textbf{82.6} & \textbf{92.0} & \textbf{83.9} \\
\midrule
\multirow{4}{*}{Spanish} & \multirow{2}{*}{FLEURS} & PhonoQ & 76.0 & 76.2 & 90.3 & 78.5 \\
 &  & PhonoQ-2.0 & \textbf{93.2} & \textbf{92.3} & \textbf{97.8} & \textbf{94.0} \\
\cmidrule(l){2-7}
 & \multirow{2}{*}{VoxPopuli} & PhonoQ & 69.1 & 58.7 & 90.0 & 69.1 \\
 &  & PhonoQ-2.0 & \textbf{90.3} & \textbf{91.3} & \textbf{96.2} & \textbf{91.6} \\
\bottomrule
\multicolumn{7}{l}{\scriptsize Man.:~Manner, Pl.:~Place, Voi.:~Voicing.}
\vspace{-4mm}
\end{tabular}
}
\end{table}

\subsection{Cross-Lingual Generalization to Unseen Languages}

Zero-shot transfer to French, Italian, and Russian shows substantial degradation for CTC-Phoneme (40.38\% PER), with mapped phonological features reaching an average 66.90\% F1. The CTC-Phoneme baseline was trained on the same four languages as PhonoQ-2.0, as well as additional Russian and French data, so the zero-shot comparison does not favor PhonoQ-2.0 in terms of language coverage. PhonoQ-2.0 improves to 73.56\% (+6.67 absolute), with gains across all languages: Italian (+10.8), French (+5.9), and Russian (+3.3). This suggests that structured phonological prediction transfers more robustly than post-hoc feature mapping under cross-lingual shift (Table~\ref{tab:external_lang_results}).


\begin{table}[t]
\centering
\caption{Unseen-language evaluation. PER (\%, $\downarrow$; CTC only) and segment-level F1 (\%, $\uparrow$). Vowel = average of height and backness. Avg~=~macro-F1 over 21 non-silence features.}
\footnotesize
\begin{tabularx}{\columnwidth}{llXXXXXX}
\hline
Lang.& System & PER & Man. & Vow. & Pl. & Voi. & Avg \\
\hline
\multirow{2}{*}{French} & \cellcolor{gray!15}CTC-Ph & \cellcolor{gray!15}39.96 & \cellcolor{gray!15}67.1 & \cellcolor{gray!15}67.9 & \cellcolor{gray!15}70.5 & \cellcolor{gray!15}90.5 & \cellcolor{gray!15}70.4 \\
& PhonoQ-2.0  & --    & 71.5 & 79.6 & 73.4 & 92.9 & 76.3 \\
\hline
\multirow{2}{*}{Italian} & \cellcolor{gray!15}CTC-Ph & \cellcolor{gray!15}26.52 & \cellcolor{gray!15}57.2 & \cellcolor{gray!15}75.7 & \cellcolor{gray!15}54.6 & \cellcolor{gray!15}88.0 & \cellcolor{gray!15}64.8 \\
 & PhonoQ-2.0  & --    & 71.5 & 82.1 & 67.6 & 92.8 & 75.6 \\
\hline
\multirow{2}{*}{Russian} & \cellcolor{gray!15}CTC-Ph & \cellcolor{gray!15}54.67 & \cellcolor{gray!15}67.9 & \cellcolor{gray!15}53.0 & \cellcolor{gray!15}66.8 & \cellcolor{gray!15}89.9 & \cellcolor{gray!15}65.5 \\
& PhonoQ-2.0  & --    & 72.7 & 55.9 & 67.7 & 94.4 & 68.8 \\
\bottomrule
\multicolumn{8}{l}{\scriptsize Man.: Manner, Vow.: Vowel, Pl.: Place, Voi.: Voicing, CTC-Ph: CTC-Phoneme.}
\end{tabularx}
\label{tab:external_lang_results}
\end{table}

\section{Per-Feature Analysis}

Segment-level F1 scores (Fig.~\ref{fig:per_feature_f1}) show that improvements extend across the entire phonological inventory rather than being concentrated in a single category. Gains are observed for all major manner classes, across vowel height and backness dimensions, and for each place contrast, while even voicing, already strong in the baseline, shows further refinement. Importantly, improvements remain substantial even when excluding affricates, the lowest-performing class in the baseline, indicating that gains are not driven by a single feature.

A complementary precision–recall analysis (Fig.~\ref{fig:per_feature_f1_scatter}) reveals that most arrows shift toward the upper-right quadrant, reflecting simultaneous gains in precision and recall rather than trade-offs. The largest movements occur for affricates, approximants, and palatal/postalveolar contrasts, where the baseline exhibits asymmetric behavior. Together, these findings indicate a systematic enhancement of the phonological representation rather than isolated corrections.

\section{Discussion \& Conclusions}

We presented PhonoQ-2.0, a multilingual frame-level phonological feature recognizer built on a pretrained XLSR wav2vec~2.0 encoder. Importantly, CTC-Phoneme is a strong phonetic baseline, achieving an average 6.80\% PER yet its post-hoc mapped phonological features fall behind PhonoQ-2.0 across all languages and conditions. This is most apparent for Spanish, where CTC-Phoneme achieves a very low 3.49\% PER yet still falls behind PhonoQ-2.0 by +7.6 F1, indicating that even highly accurate phoneme recognition does not translate directly to accurate phonological feature prediction. 
Cross-lingual transfer beyond the PhonoQ-2.0 training languages remains challenging for both systems, though PhonoQ-2.0 maintains a consistent advantage (+3.1 to +10.8 F1), suggesting that articulatory grounding provides some resilience to unseen phoneme inventories. These results demonstrate that direct phonological feature prediction is a viable and advantageous alternative to phoneme-first pipelines, with natural applications in pronunciation assessment, language learning, and low-resource speech processing.

Several limitations point to directions for future work. The 22-dimensional feature
inventory does not capture all phonological contrasts present in the evaluated languages:
French nasal vowels lack a nasality feature, Russian palatalization is absent from the feature space, and Italian geminates are not distinguished from singletons. 
Extending the feature inventory and evaluating transfer to more typologically distant languages remain important next steps.


\section{Generative AI Use Disclosure}
Generative artificial intelligence tools were used solely to assist with language editing and clarity of presentation. All research ideas, methodology, experiments, and interpretations were conceived and carried out by the authors, who take full responsibility for the originality, validity, and integrity of the work.

\bibliographystyle{IEEEtran}
\bibliography{mybib}

@inproceedings{mcauliffe17_interspeech,
  title     = {{Montreal Forced Aligner: Trainable Text-Speech Alignment Using Kaldi}},
  author    = {Michael McAuliffe and Michaela Socolof and Sarah Mihuc and Michael Wagner and Morgan Sonderegger},
  year      = {2017},
  booktitle = {{Interspeech 2017}},
  pages     = {498--502},
  doi       = {10.21437/Interspeech.2017-1386},
  issn      = {2958-1796},
}

@inproceedings{klumpp2022common,
  title={Common phone: A multilingual dataset for robust acoustic modelling},
  author={Klumpp, Philipp and Arias, Tomas and P{\'e}rez-Toro, Paula Andrea and Noeth, Elmar and Orozco-Arroyave, Juan},
  booktitle={Proceedings of the Thirteenth Language Resources and Evaluation Conference},
  pages={763--768},
  year={2022}
}

@inproceedings{guevara2020crowdsourcing,
  title={Crowdsourcing Latin American Spanish for low-resource text-to-speech},
  author={Guevara-Rukoz, Adriana and Demirsahin, Isin and He, Fei and Chu, Shan-Hui Cathy and Sarin, Supheakmungkol and Pipatsrisawat, Knot and Gutkin, Alexander and Butryna, Alena and Kjartansson, Oddur},
  booktitle={Proceedings of the Twelfth Language Resources and Evaluation Conference},
  pages={6504--6513},
  year={2020}
}

@article{garofolo1993darpa,
  title={DARPA TIMIT acoustic-phonetic continous speech corpus CD-ROM. NIST speech disc 1-1.1},
  author={Garofolo, John S and Lamel, Lori F and Fisher, William M and Fiscus, Jonathan G and Pallett, David S},
  journal={NASA STI/Recon technical report n},
  volume={93},
  pages={27403},
  year={1993}
}

@inproceedings{panayotov2015librispeech,
  title={Librispeech: an asr corpus based on public domain audio books},
  author={Panayotov, Vassil and Chen, Guoguo and Povey, Daniel and Khudanpur, Sanjeev},
  booktitle={2015 IEEE international conference on acoustics, speech and signal processing (ICASSP)},
  pages={5206--5210},
  year={2015},
  organization={IEEE}
}

@inproceedings{kath2022carina,
  title={Carina--a corpus of aligned german read speech including annotations},
  author={Kath, Hannes and Stone, Simon and Rapp, Stefan and Birkholz, Peter},
  booktitle={ICASSP 2022-2022 IEEE International Conference on Acoustics, Speech and Signal Processing (ICASSP)},
  pages={6157--6161},
  year={2022},
  organization={IEEE}
}

@inproceedings{ardila2020common,
  title={Common voice: A massively-multilingual speech corpus},
  author={Ardila, Rosana and Branson, Megan and Davis, Kelly and Kohler, Michael and Meyer, Josh and Henretty, Michael and Morais, Reuben and Saunders, Lindsay and Tyers, Francis and Weber, Gregor},
  booktitle={Proceedings of the twelfth language resources and evaluation conference},
  pages={4218--4222},
  year={2020}
}

@inproceedings{mortensen2016panphon,
  title={Panphon: A resource for mapping IPA segments to articulatory feature vectors},
  author={Mortensen, David R and Littell, Patrick and Bharadwaj, Akash and Goyal, Kartik and Dyer, Chris and Levin, Lori},
  booktitle={Proceedings of COLING 2016, the 26th international conference on computational linguistics: Technical papers},
  pages={3475--3484},
  year={2016}
}

@inproceedings{conneau21_interspeech,
  title     = {{Unsupervised Cross-Lingual Representation Learning for Speech Recognition}},
  author    = {Alexis Conneau and Alexei Baevski and Ronan Collobert and Abdelrahman Mohamed and Michael Auli},
  year      = {2021},
  booktitle = {{Interspeech 2021}},
  pages     = {2426--2430},
  doi       = {10.21437/Interspeech.2021-329},
  issn      = {2958-1796},
}

@book{arias2022analysis,
	author      = {Arias-Vergara, T.},
	title       = {{Analysis of Pathological Speech Signals}},
	booktitle   = {Studien zur Mustererkennung},
	publisher   = {Logos Verlag Berlin},
	address     = {Germany},
	year        = {2022},
	issn 		= {978-3-8325-5561-0}
}

@inproceedings{arias2023measuring,
  title={Measuring Phonological Precision in Children with Cleft Lip and Palate.},
  author={Arias-Vergara, Tom{\'a}s and Londo{\~n}o-Mora, Elizabeth and P{\'e}rez-Toro, Paula Andrea and Schuster, Maria and N{\"o}th, Elmar and Orozco-Arroyave, Juan Rafael and Maier, Andreas},
  booktitle={INTERSPEECH},
  pages={4638--4642},
  year={2023}
}

@inproceedings{arias2024contrastive,
  title={Contrastive learning approach for assessment of phonological precision in patients with tongue cancer using MRI data},
  author={Arias-Vergara, Tom{\'a}s and P{\'e}rez-Toro, Paula Andrea and Liu, Xiaofeng and Xing, Fangxu and Stone, Maureen and Zhuo, Jiachen and Prince, Jerry L and Schuster, Maria and N{\"o}th, Elmar and Woo, Jonghye and others},
  booktitle={Interspeech},
  volume={2024},
  pages={927},
  year={2024}
}

@inproceedings{liu2025audio,
  title={Audio--vision contrastive learning for phonological class recognition},
  author={Liu, Daiqi and Arias-Vergara, Tom{\'a}s and Hutter, Jana and Maier, Andreas and P{\'e}rez-Toro, Paula Andrea},
  booktitle={International Conference on Text, Speech, and Dialogue},
  pages={60--71},
  year={2025},
  organization={Springer}
}

@inproceedings{parlaspeech,
    title = "{P}arla{S}peech-{HR} - a Freely Available {ASR} Dataset for {C}roatian Bootstrapped from the {P}arla{M}int Corpus",
    author = "Ljube{\v{s}}i{\'c}, Nikola  and
      Kor{\v{z}}inek, Danijel  and
      Rupnik, Peter  and
      Jazbec, Ivo-Pavao",
    editor = "Fi{\v{s}}er, Darja  and
      Eskevich, Maria  and
      Lenardi{\v{c}}, Jakob  and
      de Jong, Franciska",
    booktitle = "Proceedings of the Workshop ParlaCLARIN III within the 13th Language Resources and Evaluation Conference",
    month = jun,
    year = "2022",
    address = "Marseille, France",
    publisher = "European Language Resources Association",
    pages = "111--116",
}

@inproceedings{parlaspeech3,
author="Kopp, Maty{\'a}{\v{s}}
and Stankov, Vladislav
and Kr{\r{u}}za, Jan Old{\v{r}}ich
and Stra{\v{n}}{\'a}k, Pavel
and Bojar, Ond{\v{r}}ej",
editor="Ek{\v{s}}tein, Kamil
and P{\'a}rtl, Franti{\v{s}}ek
and Konop{\'i}k, Miloslav",
title="ParCzech 3.0: A Large Czech Speech Corpus with Rich Metadata",
booktitle="Text, Speech, and Dialogue",
year="2021",
publisher="Springer International Publishing",
address="Cham",
pages="293--304",
isbn="978-3-030-83527-9"
}

@article{fleurs2022arxiv,
  title = {FLEURS: Few-shot Learning Evaluation of Universal Representations of Speech},
  author = {Conneau, Alexis and Ma, Min and Khanuja, Simran and Zhang, Yu and Axelrod, Vera and Dalmia, Siddharth and Riesa, Jason and Rivera, Clara and Bapna, Ankur},
  journal={arXiv preprint arXiv:2205.12446},
  year = {2022},
}

@inproceedings{wang-etal-2021-voxpopuli,
    title = "{V}ox{P}opuli: A Large-Scale Multilingual Speech Corpus for Representation Learning, Semi-Supervised Learning and Interpretation",
    author = "Wang, Changhan  and
      Riviere, Morgane  and
      Lee, Ann  and
      Wu, Anne  and
      Talnikar, Chaitanya  and
      Haziza, Daniel  and
      Williamson, Mary  and
      Pino, Juan  and
      Dupoux, Emmanuel",
    booktitle = "Proceedings of the 59th Annual Meeting of the Association for Computational Linguistics and the 11th International Joint Conference on Natural Language Processing (Volume 1: Long Papers)",
    month = aug,
    year = "2021",
    address = "Online",
    publisher = "Association for Computational Linguistics",
    pages = "993--1003",
}

@inproceedings{tadavarthy24_interspeech,
  title     = {{Phonological Feature Detection for US English using the Phonet Library}},
  author    = {Harsha Veena Tadavarthy and Austin Jones and Margaret E. L. Renwick},
  year      = {2024},
  booktitle = {{Interspeech 2024}},
  pages     = {1515--1519},
  doi       = {10.21437/Interspeech.2024-318},
  issn      = {2958-1796},
}

@inproceedings{thienpondt2025weakly,
  title={Weakly Supervised Phonological Features for Pathological Speech Analysis},
  author={Thienpondt, Jenthe and Vanderreydt, Geoffroy and Hammami, Abdessalem and Demuynck, Kris},
  booktitle={ICASSP 2025-2025 IEEE International Conference on Acoustics, Speech and Signal Processing (ICASSP)},
  pages={1--5},
  year={2025},
  organization={IEEE}
}

@inproceedings{cho2024self,
  title={Self-supervised models of speech infer universal articulatory kinematics},
  author={Cho, Cheol Jun and Mohamed, Abdelrahman and Black, Alan W and Anumanchipalli, Gopala K},
  booktitle={ICASSP 2024-2024 IEEE International Conference on Acoustics, Speech and Signal Processing (ICASSP)},
  pages={12061--12065},
  year={2024},
  organization={IEEE}
}

@inproceedings{lee2025leveraging,
  title={Leveraging IPA and articulatory features as effective inductive biases for multilingual ASR training},
  author={Lee, Jaeyoung and Mimura, Masato and Kawahara, Tatsuya},
  booktitle={ICASSP 2025-2025 IEEE International Conference on Acoustics, Speech and Signal Processing (ICASSP)},
  pages={1--5},
  year={2025},
  organization={IEEE}
}

@inproceedings{kim2025improving,
  title={Improving cross-lingual phonetic representation of low-resource languages through language similarity analysis},
  author={Kim, Minu and Jang, Kangwook and Kim, Hoirin},
  booktitle={ICASSP 2025-2025 IEEE International Conference on Acoustics, Speech and Signal Processing (ICASSP)},
  pages={1--5},
  year={2025},
  organization={IEEE}
}

@inproceedings{choi2025leveraging,
  title={Leveraging allophony in self-supervised speech models for atypical pronunciation assessment},
  author={Choi, Kwanghee and Yeo, Eunjung and Chang, Kalvin and Watanabe, Shinji and Mortensen, David R},
  booktitle={Proceedings of the 2025 Conference of the Nations of the Americas Chapter of the Association for Computational Linguistics: Human Language Technologies (Volume 1: Long Papers)},
  pages={2613--2628},
  year={2025}
}

@incollection{clements1995internal,
  title={The internal organization of speech sounds},
  author={Clements, George N and Hume, Elizabeth V},
  booktitle={The handbook of phonological theory/ed. by John A. Goldsmith},
  pages={245--306},
  year={1995},
  publisher={Blackwell}
}

@inproceedings{vasquezcorrea19_interspeech,
  title     = {{Phonet: A Tool Based on Gated Recurrent Neural Networks to Extract Phonological Posteriors from Speech}},
  author    = {J.C. Vásquez-Correa and Philipp Klumpp and Juan Rafael Orozco-Arroyave and Elmar Nöth},
  year      = {2019},
  booktitle = {{Interspeech 2019}},
  pages     = {549--553},
  doi       = {10.21437/Interspeech.2019-1405},
  issn      = {2958-1796},
}

@inproceedings{rudzicz2009phonological,
  title={Phonological features in discriminative classification of dysarthric speech},
  author={Rudzicz, Frank},
  booktitle={2009 IEEE International Conference on Acoustics, Speech and Signal Processing},
  pages={4605--4608},
  year={2009},
  organization={IEEE}
}

@article{xu2021simple,
  title={Simple and effective zero-shot cross-lingual phoneme recognition},
  author={Xu, Qiantong and Baevski, Alexei and Auli, Michael},
  journal={arXiv preprint arXiv:2109.11680},
  year={2021}
}

\end{document}